\documentclass[letterpaper]{article} 
\usepackage{aaai24}  
\usepackage{times}  
\usepackage{helvet}  
\usepackage{courier}  
\usepackage[hyphens]{url}  
\usepackage{graphicx} 
\usepackage{natbib}  
\usepackage{caption} 
\usepackage{algorithm}
\usepackage{algorithmic}
\usepackage{amsmath}
\usepackage{subfigure}
\usepackage{multirow}
\usepackage{amsfonts}
\usepackage[table]{xcolor}
\usepackage{bm}
\usepackage{comment}

\usepackage{array}
\usepackage{newfloat}
\usepackage{listings}
\DeclareCaptionStyle{ruled}{labelfont=normalfont,labelsep=colon,strut=off} 
\floatstyle{ruled}
\newfloat{listing}{tb}{lst}{}
\floatname{listing}{Listing}
\title{Dynamic Sub-graph Distillation for Robust Semi-supervised Continual Learning}
\author{
    Yan Fan,
    Yu Wang\thanks{Corresponding author},
    Pengfei Zhu,
    Qinghua Hu
}
\affiliations{
    \textsuperscript{\rm 1}Tianjin Key Lab of Machine Learning, College of Intelligence and Computing, Tianjin University, China\\

    \textsuperscript{\rm 2}Haihe Laboratory of Information Technology Application Innovation, China\\
    fyan\_0411@tju.edu.cn, wangyu\_@tju.edu.cn, zhupengfei@tju.edu.cn, huqinghua@tju.edu.cn
}

\usepackage{bibentry}

\begin{document}

\maketitle

\begin{abstract}
Continual learning (CL) has shown promising results and comparable performance to learning at once in a fully supervised manner. However, CL strategies typically require a large number of labeled samples, making their real-life deployment challenging. In this work, we focus on semi-supervised continual learning (SSCL), where the model progressively learns from partially labeled data with unknown categories. 
We provide a comprehensive analysis of SSCL and demonstrate that unreliable distributions of unlabeled data lead to unstable training and refinement of the progressing stages. This problem severely impacts the performance of SSCL. 
To address the limitations, we propose a novel approach called Dynamic Sub-Graph Distillation (DSGD) for semi-supervised continual learning, which leverages both semantic and structural information to achieve more stable knowledge distillation on unlabeled data and exhibit robustness against distribution bias. 
Firstly, we formalize a general model of structural distillation and design a dynamic graph construction for the continual learning progress. Next, we define a structure distillation vector and design a dynamic sub-graph distillation algorithm, which enables end-to-end training and adaptability to scale up tasks. The entire proposed method is adaptable to various CL methods and supervision settings.
Finally, experiments conducted on three datasets CIFAR10, CIFAR100, and ImageNet-100, with varying supervision ratios, demonstrate the effectiveness of our proposed approach in mitigating the catastrophic forgetting problem in semi-supervised continual learning scenarios. Our code is available: https://github.com/fanyan0411/DSGD. 
\end{abstract}

\section{Introduction}

Continual learning (CL) has been commonly investigated to model the realistic learning progress in evolving environments by learning new knowledge and reinforcing existing cognition. 
Numerous efforts have been made to alleviate the issue of catastrophic forgetting of learned models when new tasks are involved \cite{de2021continual}. However, these methods heavily rely on labeled data, which poses limitations in insufficient supervision scenarios, such as face recognition and video recognition \cite{wang2021ordisco}. 

\begin{figure}
	\centering
	\includegraphics[width=0.47\textwidth]{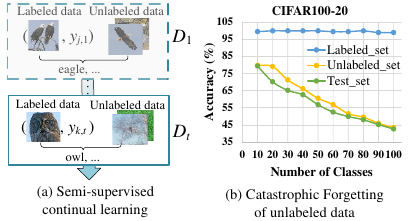}
	\caption{Challenge analysis of semi-supervised continual learning. (a) Illustration of semi-supervised continual learning.  (b) The accuracy tendency of the testing set and unlabeled training set shows a significant positive correlation.}
	\label{fig:SSCL}
\end{figure}

For the purpose of reducing reliance on annotations, semi-supervised continual learning (SSCL) is proposed and developed by exploiting massive unlabeled data to enhance the performance. An illustration of SSCL is provided in Figure \ref{fig:SSCL}(a). Nevertheless, as demonstrated in Figure \ref{fig:SSCL}(b), when all labeled samples are retrained in the subsequent tasks without any replay of unlabeled data, the decline in performance correlates with a decrease in accuracy of unlabeled training data, indicating the forgetting of unlabeled data contributes to catastrophic forgetting of the SSCL. 

To deal with the forgetting problem in SSCL, some methods employ the semi-supervised learning strategy on SSCL, such as distilling pseudo-label on unlabeled samples \cite{DistillMatch} or applying consistency loss to enhance the model discriminability \cite{CCIC}. Additionally, the online sample replaying method is utilized through training data sampled from a learned conditional generator in an online manner \cite{wang2021ordisco}. 

However, most of these strategies have paid more attention to leveraging the learned representations of unlabeled data, while unreliable distributions will have a detrimental impact on the training stability and refinement \cite{chen2022debiased, wang2021ordisco}. We provide a visualization of the negative effect of distribution bias and pseudo label errors on performance and training stability in Figure \ref{fig:distribution}. This encourages us to investigate into a more robust strategy for alleviating forgetting of SSCL. 

Considering the aforementioned observations, we propose a Dynamic Sub-graph Distillation (DSGD) designed explicitly for SSCL. We aim to enhance the robustness of the proposed SSCL method by exploring the association and structural knowledge in unlabeled data. To achieve this, we first describe the data's underlying structure through a graph representation. We then formulate a sub-graph preserving principle to model stable learning, where the graph local structure captures the underlying high-order relationships among samples. Subsequently, to ensure the scalability of our method as the learning progresses and the knowledge becomes more complex, we introduce the concept of distillation vectors based on personalized PageRank values \cite{PPR} of the dynamic graph. We finally design an efficient distillation loss that scales well with the growing complexity of the tasks. By relying less on absolute representations, our DSGD strategy can mitigate the influence of data distribution bias and pseudo-label errors, enabling more robust and effective semi-supervised continual learning. 

We follow the ORDisCo \cite{wang2021ordisco} to split commonly used CL benchmark datasets. The experiments show significant boosts in last and average accuracy across different label ratios throughout these benchmarks, with up to 60\% memory occupation savings over existing state-of-the-art approaches. In summary, our contributions are threefold:
\begin{itemize} 
	\item[(1)] We provide a systematical study of the SSCL and show that unreliable distributions of unlabeled data lead to unstable training and harmful refinement in the continual learning progress. 
	\item[(2)] We propose a novel method called Dynamic Sub-graph Distillation (DSGD) that leverages higher-order structures of association information to improve the robustness against hurtful distribution bias and mitigate the catastrophic forgetting of SSCL. 
	
	\item[(3)] Through comprehensive experiments on three commonly used benchmarks, we show that our method improves the catastrophic forgetting problem of SSCL, highlighting its robustness in various supervision scenarios and effectiveness on practical relevance.
\end{itemize} 
\section{Related Work}
\subsection{Continual Learning}
Continual learning (CL) methods can be organized into three aspects: reply-based methods, regularization-based methods, and parameter isolation methods.

Reply-based methods select representative samples for retraining when learning new concepts. For instance, iCaRL \cite{icarl} uses the approximated class means. GEM \cite{GEM} constrains new task updates to not interfere with previous tasks. ERC \cite{ERC} proposes the reservoir sampling scheme. 
Generative replay methods \cite{PR} model the distribution and generate instances for rehearsal without revisiting prior samples. 
Regularization-based methods try to employ extra regularization loss to consolidate prior knowledge during the learning process on novel data,  such as penalizing changes to essential parameters \cite{EWC} or distilling output of the previous model and the new model \cite{LWF}. 
Parameter isolation methods dedicate different model parameters to each task. DER \cite{DER} freezes previously learned extractor and expands a new backbone when facing new tasks. 
To alleviate the catastrophic expansion, some studies design a feature-boosting strategy \cite{FOSTER} or decouple the backbone at the middle layers instead of the entire network \cite{MEMO}. 
Knowledge distillation and data replay dominated the research before the presentation of DER, while dynamic networks became popular after DER \cite{pycil}. 
\subsection{Semi-supervised Learning}

\begin{figure}
	\small
	\centering
	\includegraphics[width=0.46\textwidth]{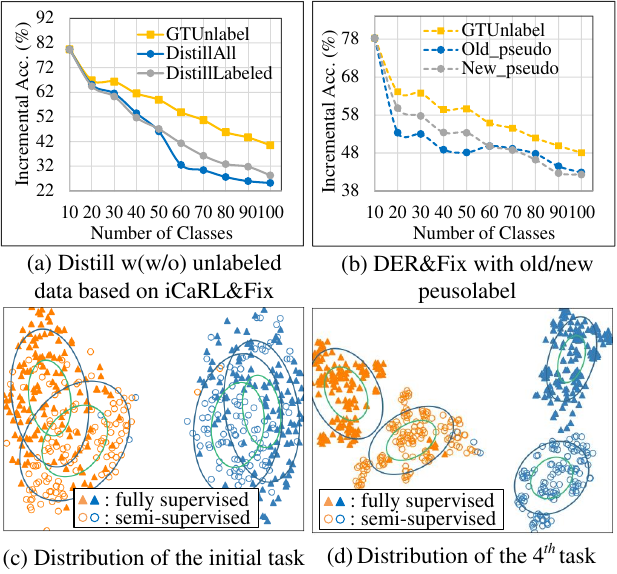}
	\caption{Baselines that combine CL and SSL methods to tackle SSCL. (a) DistillLabel: distillation on labeled data. DistillALL: apply distillation loss on the entire memory buffer. GTUnlabel: correct the pseudo-labels of unlabeled data in the memory buffer to ground truth. (b) New(Old)\_pseudo: apply the predictions of the current(previous) network as pseudo labels $\bm{p}^{\mathcal{A}}$. (c-d) The distribution bias of fully and semi-supervised settings. } 
	\label{fig:distribution}
\end{figure}

Semi-supervised learning (SSL) presents a general framework for harnessing the potential of unlabeled data. 
Pseudo-labeling methods assign the predictions of unlabeled data as pseudo labels to enlarge the training set \cite{lee2013pseudo, xie2020self}. Consistency regularization methods constrain the predictions of different augmented distributions to be close through teacher-student model interactions \cite{meanteacher} or adversarial perturbations on inputs \cite{VAT} to expand the generality boundary. 
To expand the margin with unlabeled data, FixMatch \cite{fixmatch} applies AutoAugment \cite{cubuk2019autoaugment} to build a stronger augmentation version compared to the weaker one. The impressive results of FixMatch have pushed forward more studies \cite{wang2022freematch, chen2023softmatch} for adapting to more complex situations.

Nevertheless, existing efforts of SSL do not fully account for the potential variations in the data distribution or category over time.

\subsection{Semi-supervised Continual Learning}  
Semi-supervised Continual Learning (SSCL) considers a more realistic continual task stream that only a limited number of samples are annotated. 
The success of SSL allows a general effort to exploit unlabeled data to improve performance throughout the entire stage. 
CNNL \cite{CNNL} fine-tunes its incremental learner by generating the pseudo-labels of unlabeled to enable self-training. DistillMatch \cite{DistillMatch} employs knowledge distillation with prediction consistency on unlabeled data and optimizes an out-of-distribution detector to identify task-specific representations. NNCSL \cite{NNCSL} proposes class-instance similarity distillation to preserve relation information. Pseudo Gradient Learners \cite{luo2022learning} proposes a gradient learner from labeled data to predict gradients on unlabeled data to avoid the risk of pseudo labels. 

Apart from SSL-based methods, generative replayed methods dedicate to dealing with the forgetting of SSCL. For instance, ORDisCo \cite{wang2021ordisco} continually learns a conditional GAN with a classifier from partially labeled data and replays data online. 
Meta-Consolidation \cite{brahma2021hypernetworks} extends ORDisCO to meta-learning setting scheme. 
Despite the learned representations of unlabeled data having a favor for expanding the classification boundary, the unreliable distribution will blur the boundary and hurt the refinement in the following tasks.

\section{Methods}

In this section, we begin by formulating the problem of Semi-supervised Continual Learning (SSCL). Subsequently, we systematically analyze the primary challenge associated with SSCL. To address this challenge, we propose Dynamic Sub-graph Distillation (DSGD) to mitigate the issue of catastrophic forgetting of unlabeled data. 

\subsection{Problem Formulation and Baseline}
The research of SSCL amounts to learning an ordered set of $T$ tasks that exhibit different data distribution $\mathcal{D}^t$. 
The data of each task sampled i.i.d. from $\mathcal{D}^t$ with few annotations $D^t=\{(X^t_l,Y^t_l),X^t_u\}$, where $X^t_l$ and $X^t_u$ represents the labeled and unlabeled data, respectively. In this paper, we consider class continual learning so that for any two tasks $Y^s\cap Y^t=\emptyset$.  
Each task is a specific semi-supervised learning process, which attempts to find a model $f:X^t\rightarrow Y^t$ to map both labeled samples and unlabeled samples to the target space. 
Furthermore, the learned model should perform well on the previous tasks even without experience id by involving distillation constraints. 
Therefore, the whole object can be formulated as follows:
\begin{equation}
	\begin{split}
		\mathop{\arg\min}\limits_{\theta}\sum_{t=1}^{T}&\mathbb{E}_{(x,y)\sim \mathcal{D}^{t}}[\mathcal{L}_{CE}(\bm{p},\bm{y})\\
		& +\lambda_1 \mathcal{L}_{SSL}(\bm{p}^{\mathcal{A}},\bm{p}^{\mathcal{B}}) + \lambda_2 \mathcal{L}_{CL}(\overline{\bm{z}},\bm{z})], 
	\end{split}
	\label{eq:wholeloss}
\end{equation}
where $\mathcal{L}_{CE}$ is the cross-entropy loss, $\mathcal{L}_{SSL}$ is the semi-supervised loss and  $\mathcal{L}_{CL}$ represents the continual learning loss. $\lambda_1, \lambda_2$ are corresponding weights. We utilize the representative method Fixmatch as our SSL baseline, iCaRL and DER as our CL baselines. The SSL loss encourages prediction consistency between the strong augmentation $\bm{p}^\mathcal{B}$ and the weak augmentation $\bm{p}^\mathcal{A}$ of the same image. 
Additionally, iCaRL follows knowledge distillation by compelling the new network to generate outputs $\bm{z}$ aligning with those of the old network $\overline{\bm{z}}$. DER preserves the old network by parameter consolidation. The combined baselines in our paper are denoted as iCaRL\&Fix and DER\&Fix.    

\subsection{A Systematic Study of SSCL}
As illustrated in Figure \ref{fig:SSCL}(b),  the catastrophic forgetting of unlabeled data is an essential challenge in SSCL. We then explore the capability of CL methods, iCaRL and DER, when adapting to unlabeled data.
In particular, we conduct extensive experiments on the CIFAR100, where only 20 samples are annotated per class. The training set is divided into 10 tasks, with 10 categories assigned to each task. 

We initially conducted comparative experiments with the distillation strategy presented in 
Figure \ref{fig:distribution}(a), where it can be seen that applying the distillation strategy of iCaRL on unlabeled data leads to explicit accuracy decreases. When correcting the wrong distillation term $\overline{\bm{z}}$ into ground truth, we can observe consistent improvements across tasks. The results show that conventional distillation may introduce unreliable information for unlabeled data, causing unstable distillation and detrimental impacts on the refinement of SSCL. 

Parameter isolation methods heavily rely on the annotated examples in the memory buffer to prevent catastrophic forgetting of the classifier \cite{DER}. To explore if preserving previous classification results on unlabeled data is valuable, we consider returning old predictions of unlabeled examples in the memory buffer as pseudo labels on subsequent tasks.
Therefore, we employ two kinds of $\bm{p}^{\mathcal{A}}$, the predictions of current and previous tasks, as pseudo labels for SSL loss to deal with forgetting of SSCL.  
As illustrated in Figure \ref{fig:distribution}(b), when retraining the replayed data with the previous predictions, the accuracy decreases across several tasks. The results provide further evidence of the negative effects of preventing incorrect predictions. 

To have a deeper insight into why conventional CL fails to address the catastrophic forgetting problem on SSCL, we visualize the distribution learned under fully supervised and semi-supervised settings. 
As depicted in Figure \ref{fig:distribution}(c-d), the presence of distribution bias in the semi-supervised learning task is evident when compared to the fully supervised learning setting. Furthermore, this bias continues into the subsequent tasks.  
Accordingly, directly preserving the unreliable instance-wise representations or classification results is not appropriate for unlabeled data, which undermines the potential of unlabeled data in overcoming catastrophic forgetting in SSCL. This allows us to investigate a more robust distillation strategy for SSCL.  

\begin{figure*}[t]
	\small
	\centering
	\includegraphics[width=1\textwidth]{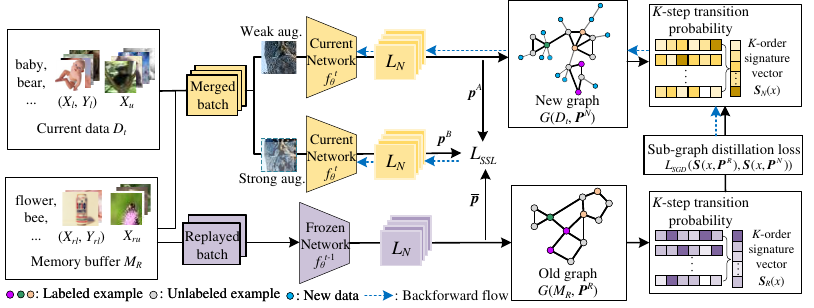}
	\caption{ The framework of the proposed method Dynamic Sub-graph Distillation. Given the merged batch of current and replayed samples, we first generate weak and strong augmentations for each image, and the semantic representations produced by the current network $f^t_\theta$. We then employ the outputs of the weak version to construct the new graph, and the corresponding old graph based on replayed samples can also be built. Through the probability matrix $\bm{P}^R$ and $\bm{P}^N$, the distillation vector, which captures the local structure information, can be used as the component of our sub-graph distillation loss. With the guides of distillation loss, the current network $f_\theta^t$ will be trained with the invariance of sub-graph structure associated with each example. }
	\label{fig:myfigure}
\end{figure*}

\subsection{Dynamic Sub-graph Distillation}
In order to address the limitations mentioned above, we introduce a novel framework that focuses on the utilization of association knowledge derived from high-order neighbors and local structure information.  
Our method is built upon the following ideas: (1) The knowledge acquired in the brain is interconnected and structural,  ensuring the knowledge evolves with fundamental structure stability. (2) Through using graph-based techniques, we build a projection from the old topology graph to the new one and preserve essential local structures. The framework is shown in Figure \ref{fig:myfigure}. 

We formalize the proposed method as follows. Given the training data $D_t$ of the new stage containing previous data $M_R$, we first construct the new topology graph $G(D_t, \bm{E}^N)$ and the topology graph of replayed data $G(M_R, \bm{E}^R)$, where $\bm{E}$ represents the adjacency matrix. We denote $\bm{S}(x, \bm{E})$ as the encoding of the node sub-structure in the graph. The learning progress is considered sub-structure preserving if there exists a mapping $\phi$ from the old knowledge based on replayed data $G(M_R, \bm{E}^R)$ to the new one $G(D_t, \bm{E}^N)$, such that $\bm{S}(x, \bm{E}^R) = \bm{S}(\phi(x), \bm{E}^N)$.
We intuitively use the identity mapping $\phi(x) = x$, and design a distillation objective to ensure its sub-structure preserving property: 
\begin{equation}
	\begin{split}
		\mathop{\arg\min}\limits_{\theta}\mathbb{E}_{(x,y)\sim M_R}[\mathcal{L}(\bm{S}(x, \bm{E}^R),\bm{S}(\phi(x), \bm{E}^N))]. 
	\end{split}
	\label{eq:objective}
\end{equation}

\textbf{Structural Similarity.}
Graph Matching is a general method to compare two structures with node connections by exploiting structural information and features. Personalized PageRank (PPR) \cite{PPR} value quantifies the connections between two vertices. The proximity of PPR values for vertices $u$ and $v$ indicates a higher likelihood of the pair $[u,v]$ being a valid match. For any vertex $u\in V$, its PPR value $\pi(s,u)$ w.r.t. source vertex $s$ is the probability that a random walk from $s$ terminates at $u$. Starting from vertex $s$, let $q^{(t)}_{su}$ be the probability that the random walk reaches vertex $u$ after $t$ steps, then $\pi(s,u)=\sum_{t=0}^{\infty}q^{(t)}_{su}$. 

Graph matching involves a predefined graph structure and aims to establish correspondences between nodes using structural similarity. However, with the model training process in SSCL, the graph structure of new tasks is often unknown, and each batch of samples varies. To ensure the consistency of the graph structure in such scenarios, it becomes essential to design an approach capable of adapting to dynamically changing graph structures. 

\textbf{Dynamic Topology Graph Construction.}
To achieve end-to-end training, we need to explore and use the structure information on batch samples in a dynamic manner. 
We first use the herding strategy to select prior exemplars without requiring class ID. 
Given the merged batch of current and replayed data, we need to build two associated knowledge graphs. In semi-supervised learning, the widely adopted manifold assumption suggests that representations of similar instances should be closer in feature space. Guided by this assumption, we build the old topology graph based 
on the cosine similarity of representations $\bm{Z} = f_\theta^{t-1}(M_R)$, where $f_\theta^{t-1}$ is the feature extractor trained on the previous task. 
The similarity then can be represented in matrix forms: 
\begin{equation}
	\bm{A} = \hat{\bm{Z}}\hat{\bm{Z}}^{T},
	\label{eq:semantic-simi}
\end{equation}
where $\hat{\bm{Z}}$ is the normalized features. 
To compute the PPR value, we 
define the probability transition matrix $\bm{P}$ by 

\begin{equation}
	\bm{P}_{ij} = \frac{\exp(\bm{A}_{ij}/\gamma)}{\sum^{|M_R|}_{i=1}{\exp(\bm{A}_{ij}/\gamma)}}.
	\label{eq:probabilitymatrix}
\end{equation}
The vector $\bm{P}_j$ represents the probability transition started from vertex $v_j$ satisfying $\sum_{i=1}^{|M_R|}\bm{P}_{ij}=1$. The parameter $\gamma$ controls the smoothness of the transition matrix. 
We then get the old topology graph $G(M_R, \bm{P}^R)$, where higher similarity between two node's representation leads to higher transition probability. 
Similarly, we define the new topology graph $G(D_t, \bm{P}^N)$ utilizing the updated embedding $\bm{Z}_t=f_\theta^t(D_t)$. 

The appearance of new tasks leads to the acquisition of new knowledge. As a result, the ability to recognize new associations and comprehend deeper structural information expands and evolves. Our graph structure is specifically designed to adapt to these dynamic processes, enabling the accommodation of evolving learning tasks. 

\textbf{Dynamic Sub-graph Distillation.} 
Based on the Equation (\ref{eq:objective}), we propose a dynamic graph structure distillation mechanism to 
combat the issue of catastrophic forgetting of unlabeled data effectively. We quantify the stability of old knowledge by evaluating the invariance of the sub-graph structure induced by each sample on the topology graph. Accordingly, we define the PPR value associated with the replayed sample as a representation of the sub-graph structure. Notably, as our adjacency matrix is fully connected, the original PPR value is toward infinity. To address this problem, we propose the $K-$order PPR value $\pi^K(s,u)=\sum_{t=0}^{K}q^{(t)}_{su}$, which represents a probability that a random walk from $s$ terminated at $u$ within $K$ steps.  
Such probabilities can be represented in matrix forms. Let $\bm{e}_s\in \mathbb{R}^{|M|*1}$ be $s^{th}$ unit vector, i.e. with 1 at the $s^{th}$ position and 0 everywhere else. Let $\bm{P}$ be the transition matrix, 
then the PPR value of vertex $u$ w.r.t. source $s$ is 
\begin{equation}
	\pi^K(s,u)=\sum_{t=0}^{K}[\bm{P}^t \cdot \bm{e}_s]_u.
	\label{eq:ppr}
\end{equation}

Given a set of starting vertex, we define the distillation vector that signifies the high-order topology structure of replayed sample $x$ on the old graph as $\bm{S}_R(x) = \{\pi^K_R(s_1,x), \pi^K_R(s_2,x), \dots, \pi^K_R(s_{|V|},x)\}$, where $s$ is the starting vertex. Similarly, the distillation vector of the same example $x$ on new graph is $\bm{S}_N(x) = \{\pi^K_N(s_1,x), \pi^K_N(s_2,x), \dots, \pi^K_N(s_{|V|},x)\}$. Intuitively, the closer the distillation vectors $\bm{S}_R(x)$ and $\bm{S}_N(x)$ are, the better the local structure is preserved. 
During the training stage, only a small size of samples could be available in each batch, so we use all the examples in the memory buffer as the starting vertex.  
Thus, we propose to define the dynamic distillation loss through the sub-graphs associated with examples.
\begin{equation}
	\begin{split}
		\mathcal{L}_{SGD} &=\mathcal{L}(\bm{S}(x, \bm{P}^R),\bm{S}(x, \bm{P}^N)) \\
		&= \mathcal{L}(\bm{S}_R(x), \bm{S}_N(x)) \\ 
		&= \sum_{s_i\in M_R}(\pi^K_R(s_i,x)-\pi^K_N(s_i,x))^2.
	\end{split}
	\label{eq:SGD_loss}
\end{equation}

Regarding the semi-supervised loss in Equation (\ref{eq:wholeloss}) of data $X_{RU}$, we design a weighted sum of the predictions of current and previous networks to ensemble the supervision samples without annotations: 
$\hat{\bm{p}}^\mathcal{A} = \alpha \overline{\bm{p}} + (1-\alpha) \bm{p}^\mathcal{A}$. 
As the learning task progresses, the predictions of examples selected for rehearsal become more reliable, so we design $\alpha$ to increase in a logistic manner $\alpha = 1/(1+\exp^{(-1-T/2)})$. 

\section{Experiments}
In this section, we compare our DSGD with other methods on benchmark datasets. Then we conduct ablation studies to assess the significance of each component and provide more insights into the effectiveness of our approach. 
\subsection{Experiment Setups}
\textbf{Datasets.} We validate our method on the widely used benchmark of class continual learning \textbf{CIFAR10} \cite{cifar10}, \textbf{CIFAR100} \cite{cifar10} and \textbf{ImageNet-100} \cite{imagenet}. 
CIFAR-10 is a dataset containing colored images classified into 10 classes, which consists of 50,000 training samples and 10,000 testing samples of size 32 * 32.
CIFAR-100 comprises 50,000 training images with 500 images per class and 10,000 testing images with 100 images per class.
ImageNet-100 is composed of 100 classes with 1300 images per class for training and 500 images per class for validation. 
ImageNet-100 resembles real-world scenes with a higher resolution of 256*256.

\textbf{Implementation Details.} 
For CIFAR10, CIFAR100, and ImageNet-100 datasets, we separately train all 10, 100, and 100 classes gradually with 2, 10 and 10 classes per stage. We use a fixed memory size of 2,000 exemplars, assigning 500 samples to labeled data and the remaining 1,500 samples to unlabeled data under sparse annotations. For the semi-supervised setting, we follow ORDisCo to allocate a small number of labels for each class and adhere to the standard experiment setup for selecting the labeled data \cite{realistic}. 
To simplify the notation, we denote the benchmark as ``dataset-(number of labels/class)". For example, CIFAR10-30 indicates CIFAR10 with 30 labeled samples per class. Please see the Appendix for more details. 

\textbf{Baseline and Metrics.} For CIFAR-10 and CIFAR-100, we employ a modified ResNet-32 \cite{resnet18} as our feature extractor, and adopt the standard ResNet-18 \cite{resnet18} as the feature extractor for ImageNet-100. We follow the Methods section and apply iCaRL\&Fix and DER\&Fix as the baselines and maintain the same architecture.

\newcolumntype{C}[1]{>{\centering\arraybackslash}p{#1}}

\begin{table*}[!h]
	\small
	\renewcommand\tabcolsep{6pt}     
	\renewcommand\arraystretch{1}        
    \begin{tabular}{|l|c|c|c|c|c|c|c|c|}
		\hline
		\multirow{2}{*}{Method} &  \multicolumn{2}{c|}{CIFAR100-20 (4\%)}& \multicolumn{2}{c|}{CIFAR100-25 (5\%)}  & \multicolumn{2}{c|}{CIFAR100-80 (16\%)} & \multicolumn{2}{c|}{CIFAR100-125 (25\%) }\\
		\cline{2-9}
	    & Avg & Last & Avg & Last   & Avg & Last & Avg & Last \\
		\hline
		iCaRL          & 26.43 & 13.92        & 28.14 & 15.29     & 36.32 & 19.10 & 44.14 & 30.73  \\
		DER            & 31.01 & 23.53        & 32.82 & 26.53     & 53.32 & 41.55 & 57.21 & 48.86 \\
		DistillMatch   & -     & -            & -    & -     & -     & 37$_{20\%}$    & - & - \\
		CCIC$_{5120}$  & -     & 29.5$_{5\%}$ & -    & 29.5  & -     & - & -     &44.30  \\
		NNCSL$_{5120}$ & 55.19 & 43.53        & 57.45  & 46.0  & 67.27     & 55.37 & 67.58 &56.40  \\
		
		\hline
		iCaRL\&Fix     & 45.75 & 23.40 & 49.83 & 31.25  & 53.46 & 32.21 & 56.87 & 41.38 \\
		\ \ + DSGD     & 52.80 & 35.47 & 53.42 & 35.95 & 57.92 & 37.81 & 58.08 & 43.14 \\
		\rowcolor{gray!20} Improvement   & \textbf{\ \ +7.05} & \textbf{\ \ +12.07} & \textbf{\ \ +3.59} & \textbf{\ \ +4.70}  & \textbf{\ \ +4.46} & \textbf{\ \ +5.60} & \textbf{\ \ +1.21} & \textbf{\ \ +1.76} \\
		\hline
		DER\&Fix        & 51.76  & 40.86 & 52.03 & 44.47  &  64.03  & 50.25  & 66.69 & 53.57 \\
		\ \ + DSGD      & 55.63  & 44.63 & 57.94 & 46.68  &  65.48  & 55.40  & 69.14 & 58.50 \\
		\rowcolor{gray!20} Improvement   & \textbf{\ \ +3.87} & \textbf{\ \ +3.77} & \textbf{\ \ +5.91} & \textbf{\ \ +2.21}  & \textbf{\ \ +1.45 } & \textbf{\ \ +5.15} & \textbf{\ \ +2.45} & \textbf{\ \ +4.93} \\
		\hline
		JointTrain     &  -  & 42.27 & - & 49.97 & - & 58.11 & - & 64.08 \\
		\hline
	\end{tabular}
	\centering	
	\caption{Average incremental accuracy and the last incremental accuracy. NNSCL$_{5120}$ is the state-of-art method that conducts experiments on various label ratios, where the subscript 5120 represents the memory buffer size. The percentage values in brackets and subscripts are label ratios. Improvement represents the progress of our method DSGD compared to corresponding baselines. JointTrain means learning the entire dataset at once. }
	\label{tab:cifar100}
\end{table*}

Following previous research on continual learning \cite{DER}, we compare the top-1 average incremental accuracy: 
\begin{equation}
	A = \frac{1}{T}\sum_{t=1}^{t}{A_t}, 
\end{equation}
where $A_t$ is the incremental accuracy on the task $t$ and is defined by $A_t=\frac{1}{t}\sum_{i=1}^{t}a_{t,i}$, where $a_{t,i}$ is the accuracy on the test set of the $i^{th}$ task after learning the $t^{th}$ task. 

\subsection{Quantitative Results} 
\begin{figure}
	
	\centering
	\includegraphics[width=0.478\textwidth]{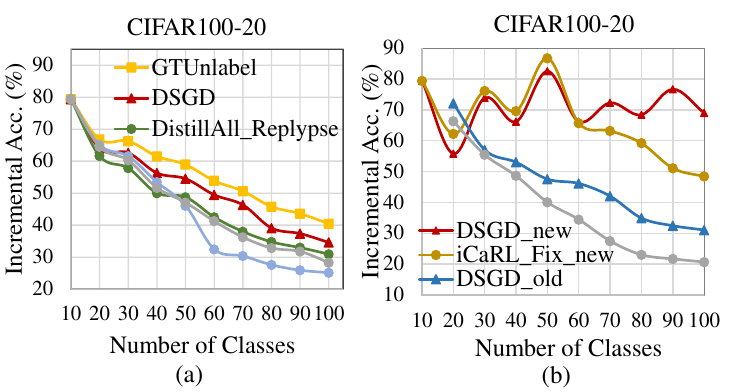}
	\caption{Accuracy of all learned tasks, all old tasks, and the new task on the CIFAR100-20 benchmark. (a) The accuracy of all learned tasks on different strategies. DistillLabeled, DistillAll represent that using representation distillation on labeled and all samples of memory, respectively; DistillAll\_Replypse means applying the previous prediction as pseudo labels based on DistillAll. (b) Accuracy of new and old tasks on our method DSGD and iCaRL\&Fix.}
	\label{fig:cifar100}
\end{figure}

\textbf{CIFAR100.} We present the performance on CIFAR100 of our method DSGD and the two baselines (iCaRL\&Fix and DER\&Fix) under four labels ratios: 4\%, 5\%, 16\% and 25\%, as shown in Table \ref{tab:cifar100} and Figure \ref{fig:cifar100}. We first validate the efficiency of combining CL and SSL directly by comparing iCaRL with iCaRL\&Fix and DER with DER\&Fix in Table \ref{tab:cifar100}. 
Nevertheless, the catastrophic forgetting of unlabeled data remains, disrupting the model's ability to retain learned knowledge in SSCL, as illustrated in Figure \ref{fig:distribution}(a-b). 
Through the following analysis, we demonstrate that DSGD effectively mitigates catastrophic forgetting of unlabeled data.  

\textbf{Our method exhibits outstanding performance with fewer annotations and lower memory buffer size. }
As shown in Table \ref{tab:cifar100}, improvements on two baselines highlight the effectiveness of our method on robust semi-supervised continual learning. Especially in scenarios with sparse labels, such as only 20 samples annotated, DSGD can remarkably increase the base model iCaRL\&Fix by 7.05\% and 12.07\% in average incremental accuracy and the last incremental accuracy, respectively. Moreover, compared to existing SSCL strategies, our method also shows superior performance. Specifically, our method surpasses NNCSL by 1.1\%, 0.68\% and 2.1\% under label ratios of 4\%, 5\%, and 25\%, respectively, while reducing the memory buffer by 60\%. 
In addition, DSGD based on iCaRL and DER significantly outperforms DistillMatch with 16\% annotations by 0.81\% and 18.4\%, and reduces 20\% annotations. DistillMatch requires all unlabeled data available, which is challenging in limited storage scenarios. Our methods only require revisiting fewer examples and are adaptable to limited storage scenarios. 

\textbf{DSGD effectively mitigates the negative effects of distribution bias.}
As shown in Figure \ref{fig:cifar100}(a), by following Figure \ref{fig:distribution}(a) and omitting the backbone iCaRL\&Fix for clarity, the results indicate that directly distilling previous logits or the pseudo labels on unlabeled data is not effective in refining the unlabeled data for subsequent processes.
The issue of distribution bias on labeled data and unreliable pseudo labels contributes to this inefficacy. 
In contrast,  the incremental accuracy improvements achieved by DSGD highlight its effectiveness in addressing this flaw by avoiding to use the distributions and, instead, exploring association information. The results in Figure \ref{fig:cifar100}(b) illustrate that DSGD notably enhances the learning of previous data without interrupting the adaptation to new tasks, where the accuracy on all old tasks evaluates the performance of refining previous data. 

\textbf{CIFAR10.}
Table \ref{tab:cifar1030&150} summarizes the experimental results for the CIFAR10-30 and CIFAR10-150 benchmarks. In the setting of only 30 annotated samples, our method based on iCaRL\&Fix surpasses the base models by 31.65\% and 45.7\% points in average accuracy and last accuracy, respectively. Even with stronger backbone DER\&Fix, our strategy achieve 8.33\% and 11.18\% accuracy improvement.
Compared to existing SSCL methods, our method exceeds CCIC by 21.21\% at 0.6\% label ratio with less data replayed, and surpasses ORDisCo by 13.78\% under 3\% labeling ratio, while ORDisCo suffers from complexity and computation cost and our method is more economical and effective. 

\textbf{Our method is flexible in adapting dimension changes and exhibits robustness against hurtful distribution bias even in severe scarce supervision scenarios.} It is usually challenging to distill knowledge in expandable CL methods, such as DER, due to dimension mismatch. Our method DSGD is capable of satisfying dimension alignment problems, so it can be integrated into different continual learning methods and achieve explicit improvements.
\begin{table}
	\small
	\centering
	\begin{tabular}{|l|c|c|c|c|}
		\hline
		\multirow{2}{*}{Method} & \multicolumn{2}{c}{CIFAR10-30 (0.6\%)}  & \multicolumn{2}{|c|}{CIFAR10-150 (3\%)} \\
		\cline{2-5}
		& Avg & Last  & Avg & Last \\
		\hline
		iCaRL           &  34.16 &  21.84 &  60.86 & 53.65 \\
		DER             &  40.41 &  31.48 &  64.77 & 61.06 \\
		CCIC$_{5120}$           & - & 55.2$_{0.8\%}$ & - & 74.3$_{5\%}$ \\
		ORDisCo         & - & - & 74.77 & 65.91 \\
		\hline
		iCaRL\&Fix          & 45.98 & 30.71 & 78.36 & 69.08 \\
		\ \ + DSGD            & 77.33 & 76.41 & 84.14 & 79.69 \\
		\rowcolor{gray!20} Improvement     & \textbf{\ \ +31.35}  &  \textbf{\ \  +45.70} &  \textbf{\ \ +5.78} &  \textbf{\ \ +10.61} \\
		\hline
		DER\&Fix           & 66.71 & 61.41 & 81.10 & 77.00 \\
		\ \ + DSGD             & 75.04 & 72.59 & 83.08 & 79.39 \\
		\rowcolor{gray!20} Improvement  &  \textbf{\ \ +8.33} &  \textbf{\ \ +11.18} &  \textbf{\ \  +1.98} &  \textbf{\ \  +2.39} \\ 
		\hline
		JointTrain    & -  &  81.64  & - & 87.76 \\
		\hline
	\end{tabular}
	\caption{Average and the last accuracy on CIFAR10.}
	\label{tab:cifar1030&150}
\end{table}

\textbf{ImageNet-100.}
We also validate the proposed methods on a higher resolution dataset ImageNet-100, where the number of annotations is 13 and 100 per class. Table \ref{tab:imagenet100} summarizes the experimental results. 
The entire results illustrate that our method is also efficacious in large-scale continual learning. Our method still outperforms the conventional CL by a large margin in 100 annotation settings, showing that our method is capable of mitigating catastrophic forgetting of unlabeled data.  Nevertheless, as shown in the JointTrain results, there is a striking gap between JointTrain and semi-supervised continual learning, indicating that although with the improvements of DSGD, the catastrophic forgetting is severe on SSCL in more realistic applications and further research should be devoted to this field. 

\begin{table}
	\small

	\begin{tabular}{|l|c|c|c|c|}
		\hline
		\multirow{2}{*}{Method} & \multicolumn{2}{c|}{ImageNet-100-13}  & \multicolumn{2}{|c|}{ImageNet-100-100} \\
		\cline{2-5}
		& Avg & Last  & Avg & Last \\
		\hline
		iCaRL     &  19.89 & 12.88 & 30.78 & 16.68 \\
		\hline
		iCaRL\&Fix      &  26.37 & 15.58 &  37.49 & 21.02  \\
		
		\ \  + DSGD   & 28.35 & 19.14 & 50.53 & 32.10 \\
		\rowcolor{gray!20}Improvement   &  \textbf{\ \ + 1.98}  & \textbf{\ \ + 3.56} & \textbf{\ \ +13.04} & \textbf{\ \ +11.8} \\
		\hline
		JointTrain    & -  &  43.72  & - & 71.56 \\
		\hline
	\end{tabular}
	\centering
	\caption{Average and the last accuracy on ImageNet-100.}
	\label{tab:imagenet100}
\end{table}

\subsection{Ablation Study and Parameter Analysis}
\begin{table}
	\small
	\begin{tabular}{|l|c|c|c|}
		
		\hline
		Method & {100-5}  &{100-20} & {100-80} \\
		
		\hline
		iCaRL\&Fix      & 31.79  & 45.75  & 53.46 \\
		iCaRL\&Fix + DSGD & 35.91  & 48.44  & 54.40 \\
		iCaRL\&Fix + PseDis   & 35.42  & 47.67  & 53.71 \\
		iCaRL\&Fix + PseDis + DSDG   & 39.85 &  52.80 & 57.92  \\
		\hline
	\end{tabular}
	\centering
	\caption{Ablations study of the proposed method DSGD. }
	\label{tab:ablation}
\end{table}
To validate the effectiveness of the proposed strategies, we conduct an ablation study on the CIFAR100 dataset with baseline iCaRL\&Fix on three different semi-supervised settings: CIFAR100-5, CIFAR100-20, and CIFAR100-80. The performance comparison is shown in Table \ref{tab:ablation}, where the DSGD means the sub-graph knowledge distillation. PseDis means utilizing pseudo labels of the previous model as logits distillation targets. The reported average accuracy across three settings can reflect the robustness of the model. 

The ablation study reveals that the dynamic sub-graph knowledge distillation can significantly improve the accuracy throughout the entire continual learning stage. This progress is explicit when the annotations are scarce, leading to a notable 4.3\% increase in average accuracy for CIFAR100-5 dataset. Additionally, the graph structure distillation can complement the logits distillation, showcasing that our methods can work in conjunction with other continual learning methods. This highlights the adaptability and effectiveness of our approach in SSCL scenarios. 

\textbf{Parameter Analysis.}
\begin{figure}
	\centering
	\includegraphics[width=0.472\textwidth]{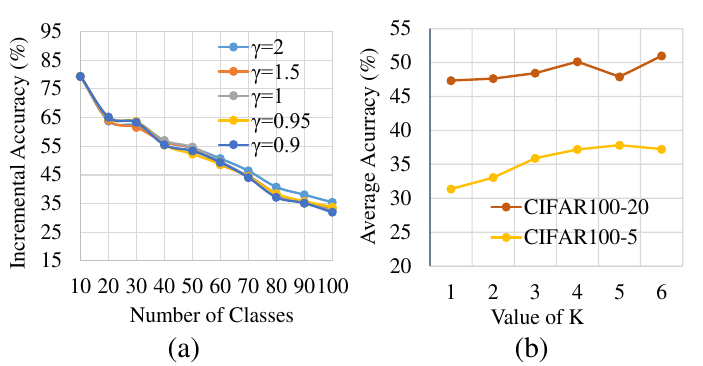}
	\caption{Robustness Testing. (a) Performance under different values of hyperparameter $\gamma$. (b) Performance with different values of K. Both of them are evaluated on CIFAR100.}
	\label{fig:parameter_analysis}
\end{figure}
To verify the robustness of DSGD, we conduct experiments on CIFAR100-20 with different hyper-parameters $\gamma \in \{0.9, 0.95, 1, 1.5, 2\}$ in dynamic topology graph construction. The results are presented in Figure \ref{fig:parameter_analysis}(a). It is evident that the performance changes are minimal across different values of $\gamma$. 

In Figure \ref{fig:parameter_analysis}(b), we gradually increase the value of $K$ in the distillation vector from 1 to 6 and record the performance of CIFAR100. In CIFAR100-20, The average accuracy increases from 47.34\% to 50.98\% as the $K$ change from 1 to 6. Similarly, the average accuracy rises from 31.37\% to 37.25\% in CIFAR100-5, indicating that our method can effectively make full use of association information. These results show that sub-graph distillation is capable of mitigating the negative effect of distribution bias of unlabeled data. 

\section{Conclusion}
Tremendous unlabeled data has the potential to improve the generalizability of continual learning significantly. However, the issue of catastrophic forgetting on unlabeled data has an impact on the performance of learned tasks. 
To address the limitations, we proposed a novel approach called Dynamic Sub-graph Distillation for robust semi-supervised continual learning, which leverages high-order structural information for more stable knowledge distillation on unlabeled data. We designed a dynamic sub-graph distillation algorithm, which enables end-to-end training and adaptability to scale up tasks. 
Experimental evaluations conducted on three datasets: CIFAR10, CIFAR100, and ImageNet-100, with varying supervision ratios, demonstrated the effectiveness of our proposed approach in mitigating the catastrophic forgetting issue in semi-supervised continual learning.

\section{Acknowledgments}
This work was supported in part by the National Key R\&D Program of China under Grant 2022ZD0116500 and in part by the National Natural Science Foundation of China under Grants 62106174, 62222608, 62266035, 61925602, U23B2049, and 62076179.

\bibliography{aaai24}

\end{document}